\documentclass[sigconf]{acmart}

\usepackage{soul}
\usepackage{color, xcolor}
\usepackage[linesnumbered,ruled,vlined]{algorithm2e}
\usepackage{ragged2e} 
\usepackage{booktabs,makecell, multirow, tabularx}
\usepackage{algpseudocode}
\usepackage[utf8]{inputenc}
\usepackage{utfsym}
\usepackage{stfloats}
\usepackage{soul}

\soulregister{\cite}7 
\soulregister{\st}7 

\AtBeginDocument{%
  }

\renewcommand\footnotetextcopyrightpermission[1]{}
\SetCommentSty{mycommfont}

\SetKwInput{KwInput}{Input}                
\SetKwInput{KwOutput}{Output}              

\begin{document}
\begin{sloppypar}

\title{RFDforFin: Robust Deep Forgery Detection for GAN-generated Fingerprint Images}

%
\author{Hui Miao}
\affiliation{%
  \institution{School of Computer Science and Engineering, Beihang University}
  \city{Beijing}
  \country{China}}
\email{huimiao@buaa.edu.cn}

\author{Yuanfang Guo}
\authornote{Corresponding author.}
\affiliation{%
\institution{School of Computer Science and Engineering, Beihang University}
  \city{Beijing}
  \country{China}}
  \email{andyguo@buaa.edu.cn}

\author{Yunhong Wang}
  \affiliation{%
  \institution{School of Computer Science and Engineering, Beihang University}
    \city{Beijing}
    \country{China}}
    \email{yhwang@buaa.edu.cn}

%

\begin{abstract}
With the rapid development of the image generation technologies, the malicious abuses of the GAN-generated fingerprint images poses a significant threat to the public safety in certain circumstances.
Although the existing universal deep forgery detection approach can be applied to detect the fake fingerprint images, they are easily attacked and have poor robustness. 
Meanwhile, there is no specifically designed deep forgery detection method for fingerprint images.
In this paper, we propose the first deep forgery detection approach for fingerprint images, which combines unique ridge features of fingerprint and generation artifacts of the GAN-generated images, to the best of our knowledge.
Specifically, we firstly construct a ridge stream, which exploits the grayscale variations along the ridges to extract unique fingerprint-specific features.
Then, we construct a generation artifact stream, in which the FFT-based spectrums of the input fingerprint images are exploited, to extract more robust generation artifact features. 
At last, the unique ridge features and generation artifact features are fused for binary classification (\textit{i.e.}, real or fake).
Comprehensive experiments demonstrate that our proposed approach is effective and robust with low complexities.
\end{abstract}



\maketitle

\begin{figure}
  \centering
  \includegraphics[width=1.0\linewidth]{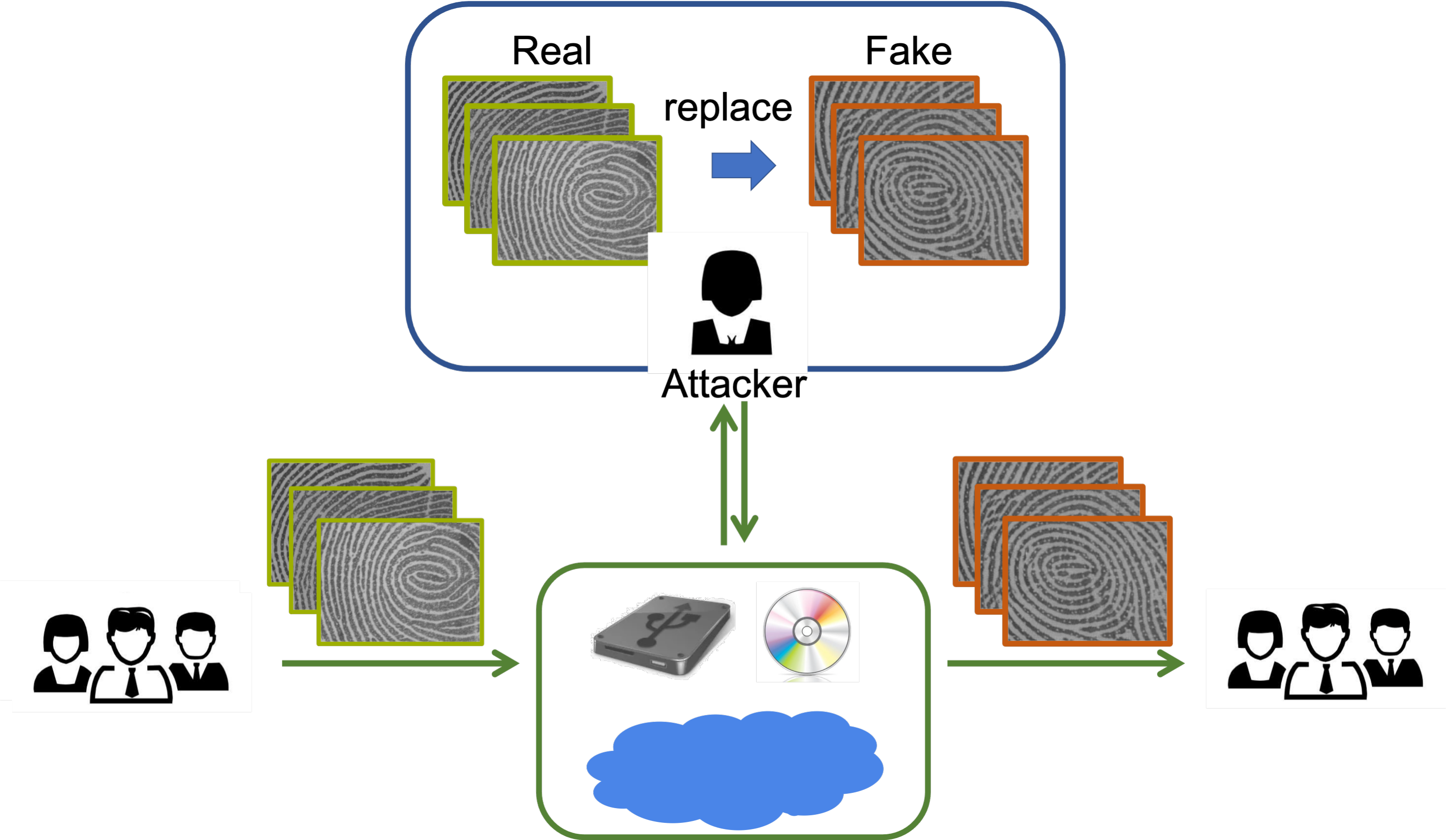}
  \caption{In police department, banks, or other institutions, the collected fingerprint data needs to be disseminated via the network or storage media~(\textit{e.g.}, disk and CD) and is vulnerable to the attacks during this process.}
  \label{fig:motivation}
  \Description{}
\end{figure}

\section{Introduction}

Deep learning based image synthesis technologies, such as Generative Adversarial Networks (GANs)~\cite{goodfellow2020generative}, has received significant attentions in the past few years. 
With the help of these techniques~\cite{zhu2017unpaired,karras2017progressive,choi2018stargan,karras2021alias}, computers can generate photorealistic images which are difficult to be distinguished from the real images.
The malicious abuses of these fake images may induce a series of security problems. In particular, fingerprint images have been widely utilized for identity authentication/verification in various applications, such as mobile payment, immigration inspection, forensics, \textit{etc.}
There exists several deep generative algorithms~\cite{bontrager2018deepmasterprints,wyzykowski2021level,engelsma2022printsgan} for generating realistic fingerprint images, which may become threats to the public safety in certain circumstances. 
As shown in Fig.~\ref{fig:motivation}, in many cases, the collected fingerprint data needs to be disseminated via the network or storage media (\textit{e.g.}, certain fingerprint data will be uploaded and stored in the data center at specific places, such as police department, bank, \textit{etc.}) If attacker maliciously replaces the collected fingerprint with a fake fingerprint in this process, the replaced fake fingerprint may induce severe consequences.
Therefore, it is essential to develop advanced deep forgery detection methods to detect these deep generated fingerprint images.

Many solutions~\cite{wang2020cnn,frank2020leveraging,liu2022detecting}, which are proposed to reduce the potential risks of general GAN-generated fake images, can be utilized to detect the GAN-generated fingerprint images in ideal situation.
~\cite{wang2020cnn} regards the deep forgery detection task as a general image classification task by directly utilizing classic backbone networks (\textit{e.g.}, ResNet [21]), whose mechanism has not focused on useful information, such as the artifacts of the generated images.
\cite{frank2020leveraging} attempts to model the artifacts, which are arose from the upsampling operations, in the frequency domain. 
\cite{liu2022detecting} extracts and learns the discrepancy of image noises, \textit{i.e.}, high frequency information, between the real and GAN-generated data. 

Unfortunately, these general solutions may not be suitable for detecting GAN-generated fingerprint images in real scenarios, which desire both the robustness to existing anti-forensic method and low complexity. 
For example, ~\cite{frank2020leveraging} is only sensitive to the spectral artifacts, which can easily be mitigated by the recent anti-forensic method~\cite{dong2022think}. 
Since~\cite{wang2020cnn,liu2022detecting} explicitly or implicitly utilize high-frequency cues and are both designed based on ResNet-50~\cite{he2016deep}, they possess a relatively large number of parameters and can also be easily disturbed by~\cite{dong2022think}.
Meanwhile, traditional fingerprint liveness detection methods~\cite{park2019presentation,chugh2020fingerprint,goicoechea2019low,liu2019high} only focus on coping with presentation attacks, which construct artificial fingers~\cite{matsumoto2002impact} to circumvent the fingerprint recognition system.
To model the physiological differences between the artificial and real fingers, existing liveness detection methods usually rely on collecting the pulse rate, skin odor, finger elasticity, \textit{etc.}, from the sensor of the fingerprint recognition system.
Unfortunately, the replacement attack, \textit{i.e.}, replacing the real fingerprint images with the GAN-generated fingerprint images, usually happens after the fingerprint collection step and the fake fingerprint images are generated without the defects of artificial fingers. 
Besides, the GAN-generated fingerprint images also exist some generation artifacts, which are induced by specific processing operations in GANs and does not exist in the captured impressions.
Under such circumstance, the existing fingerprint liveness detection methods can hardly be directly applied to detect the GAN-generated fingerprint images.

To efficiently and effectively identify the GAN-generated fingerprint images with decent robustness against the existing anti-forensic method~\cite{dong2022think},
in this paper, we propose a \textbf{R}obust deep \textbf{F}orgery \textbf{D}etection method \textbf{for} GAN-generated \textbf{Fin}gerprint images (RFDforFin). 
To take full advantage of the fingerprint characteristics and the generation artifacts of the fake fingerprint images, we construct a lightweight yet robust two-stream neural network, by exploiting the unique features of the fingerprint images and generation artifacts in frequency domain.
Considering the impact of sweat on grayscale variations along the ridges of fingerprints~\cite{derakhshani2003determination}, we construct a special ridge stream which utilizes this unique fingerprint characteristic.
In the generation artifact stream, inspired by~\cite{frank2020leveraging}, which discovers the obvious generation artifacts in the Discrete Cosine Transform (DCT) frequency domain, we transform the input fingerprint image into different frequency domains, analyze the generation artifacts in these spectrum, and construct a simple yet effective convolutional neural network to learn more robust generation artifacts between the generated and real images from the FFT frequency spectrum.
To simultaneously utilize the features extracted from the ridges and generation artifacts in the final prediction, we build a simple yet effective fusion module.
Since typical CNNs can learn certain frequency information from the input image or its frequency transformed spectrum, by utilizing the unique 1D ridge features jointly with the 2D generation artifact features, our method can avoid overfitting to certain frequency information, which can improve the robustness of our method.

Our main contributions are summarized as follows:
\begin{itemize}
\item We propose the first deep forgery detection method for GAN-generated fingerprint images, by jointly exploiting the unique 1D ridge features and 2D generation artifact features via a lightweight two-stream neural network to ensure the robustness and efficiency of the proposed work.
\item We propose to exploit fingerprint-related characteristic and construct a ridge stream, which exploits the grayscale variations along the ridges. With this ridge stream, our method can avoid overfitting to certain frequency information, which can be easily interfered by existing anti-forensic method~\cite{dong2022think} and thus improves the overall robustness.
\item We analyze the frequency spectrum of the real and generated fingerprint images and construct a simple yet effective generation artifact stream, \textit{i.e.}, a shallow convolutional neural network, to extract frequency-domain inconsistencies.
\item Comprehensive experiments demonstrate that our method is effective, efficient, and robust to the anti-forensic method~\cite{dong2022think}.
\end{itemize}

\section{Related Work}

\subsection{Fingerprint Image Synthesis}
Several recent methods have been proposed for generating realistic fingerprint images automatically~\cite{engelsma2022printsgan,grosz2022spoofgan,bontrager2018deepmasterprints,sams2022hq,bamoriya2022dsb,wyzykowski2021level,mistry2020fingerprint}.
\cite{bamoriya2022dsb} and~\cite{mistry2020fingerprint} combine a convolution autoencoder (CAE) and a GAN-based method (\textit{e.g.}, DCGAN~\cite{radford2015unsupervised}, WGAN~\cite{gulrajani2017improved}) directly.
\cite{bontrager2018deepmasterprints} presents a GAN-based pipeline followed by a stochastic search algorithm over the latent variable space, to search for suitable latent variable and generate DeepMasterPrints which are synthetic fingerprints and can be matched against a large number of fingerprints.
These techniques can be considered as single-staged architectures, whose input is a random vector and output is a generated image.

To make the synthetic results more realistic and obtain multiple impressions for a virtual identity, many multi-staged methods~\cite{engelsma2022printsgan,grosz2022spoofgan,sams2022hq,wyzykowski2021level} firstly synthesize a binary masterprint which defines a ridge structure and represents a new identity. 
After non-linear distortion and cropping, which simulate various pressures and different contact regions between the finger and the fingerprint sensor, the distorted masterprint is fed into another generative neural network to generate a realistic fingerprint image.
The differences between these multi-staged methods lie in the way of the masterprint generation scheme, the distortion simulation method, and the final impression (fingerprint) generation approach.
\cite{engelsma2022printsgan} utilizes the GAN-based methods (\textit{e.g.}, BigGAN~\cite{brock2018large}) in these three steps.
\cite{grosz2022spoofgan} attempts to employ conventional rotation instead of the deep learning method in the distortion module.
\cite{sams2022hq} adopts StyleGAN2~\cite{karras2020analyzing} to synthesize the fingerprint skeletons and CycleGAN~\cite{zhu2017unpaired} to generate photorealistic fingerprint images.
\cite{wyzykowski2021level} also proposes a CycleGAN-based L3 synthetic fingerprint generation (L3-SF) approach, which firstly uses traditional methods (\textit{e.g.}, Anguli~\cite{ansari2011generation}, Gabor filter) to generate masterprints with pores and scratch.
Besides,~\cite{wyzykowski2021level} releases a synthetic fingerprint database.

\subsection{Deep Forgery Detection}
Researchers have proposed a series of deep forgery detection approaches to detect the generated images~\cite{bammey2020adaptive,li2020face,liu2020global,wang2020cnn,li2021frequency,liu2021spatial,luo2021generalizing,chen2022self,khalid2020oc,frank2020leveraging,liu2022detecting}. 
\cite{liu2020global} reveals that the texture statistics is an important detection clue and global texture features can improve the robustness of face forgery detection. 
\cite{frank2020leveraging} observes that GAN-generated images contain distinguishable artifacts in the frequency domain due to the upsampling operations in the generator of GAN architectures, and these artifacts can be utilized for forgery detection.
\cite{liu2022detecting} attempts to explore the discrepancies in Learned Noise Patterns (LNP) to detect forgery.
\cite{liu2021spatial} concatenates spatial images and phase spectrums to emphasize the artifacts and proposes a Spatial-Phase Shallow Learning method.
\cite{li2021frequency} captures frequency-aware features and merge them with the extracted spatial domain features for the final prediction.
\cite{wang2020cnn} attempts to use data augmentations, including common pre-processing and post-processing operations, to train a ResNet-based universal detector.
\cite{khalid2020oc} adopts an one-class Variational AutoEncoder and tackles the deep facial forgery detection problem from the perspective of anomaly detection.

The majority of the above methods~\cite{liu2020global,liu2021spatial,khalid2020oc,li2021frequency,chen2022self,li2020face} are designed for deep facial forgery detection, \textit{a.k.a.} deepfake detection, which are specially designed for facial images. 
Only few of the existing methods (\textit{e.g.},~\cite{frank2020leveraging,wang2020cnn,liu2022detecting}), which are designed to detect general GAN-generated images, can be directly applied to detect the GAN-generated fingerprint images.
However,~\cite{dong2022think} reveals that the frequency artifacts are actually not reliable and proposes a pipeline to mitigate the frequency artifacts, which can actually be regarded as an anti-forensic method.
Specifically,~\cite{dong2022think} conducts a series of experiments on different types of images, including the fingerprint images, and demonstrates that the current universal GAN-generated image detectors~\cite{frank2020leveraging,wang2020cnn}, which can be applied to detect the GAN-generated fingerprint images, are not quite robust against~\cite{dong2022think}. 
Therefore, our proposed method models the fingerprint specific features to compensate the robustness of the generation artifact features from the frequency domain.

\begin{figure*}[htp]
  \centering
  \includegraphics[width=0.9\linewidth]{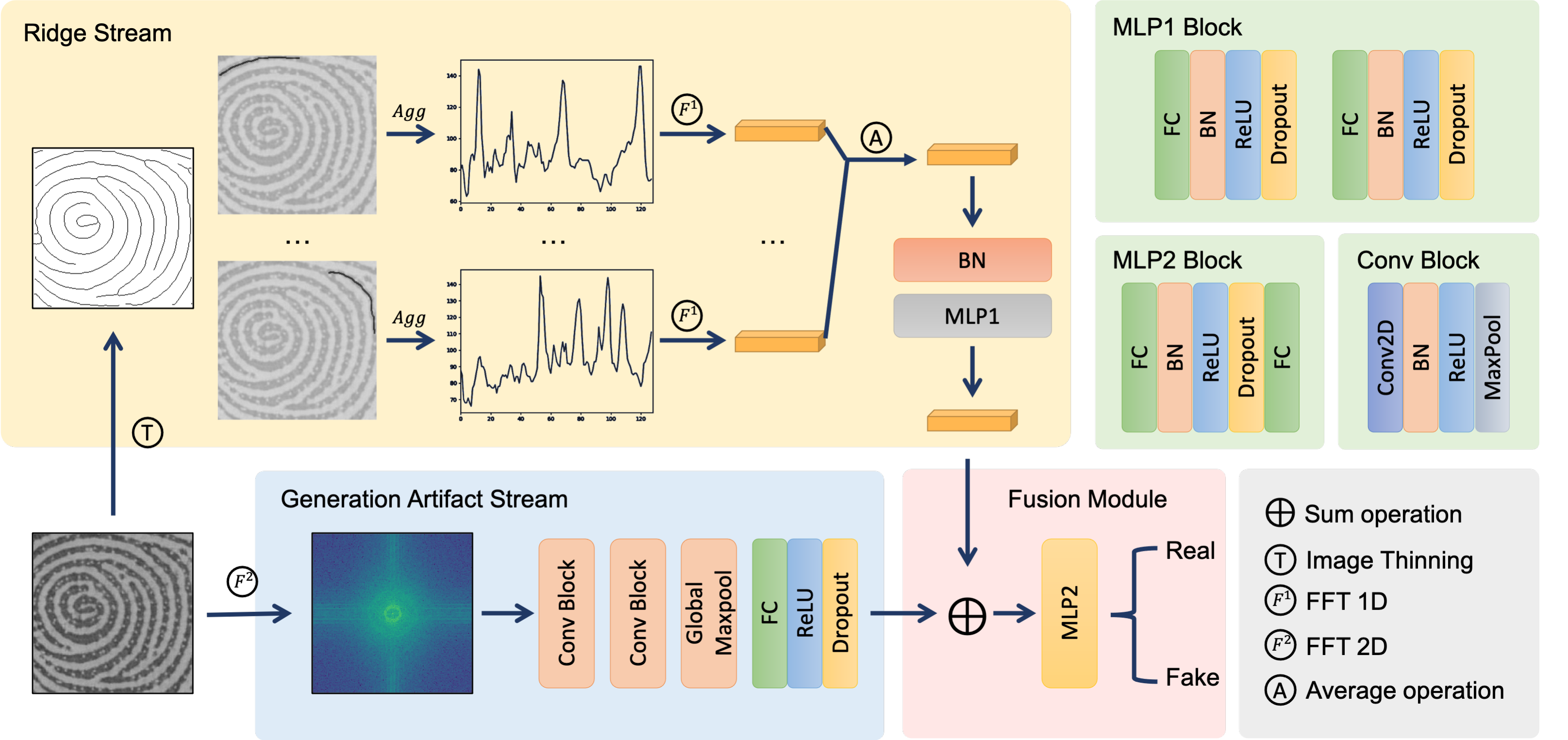}
  \caption{Our proposed method consists of a ridge stream and a generation artifact stream. In the ridge stream, we exploit the grayscale variations along the ridges in ridge feature extraction. 
  In the generation artifact stream, we construct a simple yet effective convolutional neural network to extract features from the FFT-based magnitude spectrum. At last, we fuse the features of the two streams for the final prediction.}
  \label{fig:pipeline}
  \Description{pipeline.}
\end{figure*}

\subsection{Fingerprint Liveness Detection}
Fingerprint liveness detection~\cite{park2019presentation,chugh2020fingerprint,goicoechea2019low,liu2019high,uliyan2020anti} is developed for fingerprint presentation attacks~\cite{casula2021livdet,orru2019livdet}. 
These detection methods usually focus on identifying the artifacts of the artificial fingers and thus can be classified into hardware-based techniques~\cite{goicoechea2019low,liu2019high} and software-based techniques~\cite{uliyan2020anti,chugh2020fingerprint,park2019presentation}.
The hardware-based methods mainly emphasizes the artifacts of the artificial fingers during the collection process, such as odor, surface differences, liveness signals.
On the other hand, the software-based methods consider the differences in the captured images, such as pore pattern, ridge frequencies. 
For example,~\cite{derakhshani2003determination} utilizes the perspiration pattern and combines the static ridge feature with the dynamic temporal feature during the collection process to predict the vitality of the fingerprint.
Although the software-based methods can also process the captured results, there is no collection step in our task and the generated fingerprint images exist some artifacts, which does not exist in the captured images, thus the liveness detection methods cannot be directly applied.

Since the perspiration pattern in the fingerprint images is a special fingerprint characteristic and the corresponding static ridge feature in~\cite{derakhshani2003determination} can be easily integrated into deep neural networks, we exploit this ridge feature, \textit{i.e.}, the grayscale variations along the ridges, and the GAN-generated artifacts to detect the GAN-generated fingerprint images.

\section{The Proposed Method}

In this paper, we propose the first deep forgery detection framework, which jointly exploits the unique ridge feature and generation artifacts, for fingerprint images.
As shown in Fig.~\ref{fig:pipeline}, to utilize the unique features of fingerprint images, we construct a special ridge stream, which explicitly models the grayscale variations along the ridges of fingerprint.
To better exploits the generated artifacts of the fake fingerprint images, we design a generation artifact stream, which identifies the artifacts from a FFT-based frequency spectrum, by constructing a simple yet effective convolutional neural network.
At last, we fuse these features from two streams for the final prediction.

\begin{figure}[t]
  \centering
  \includegraphics[width=1\linewidth]{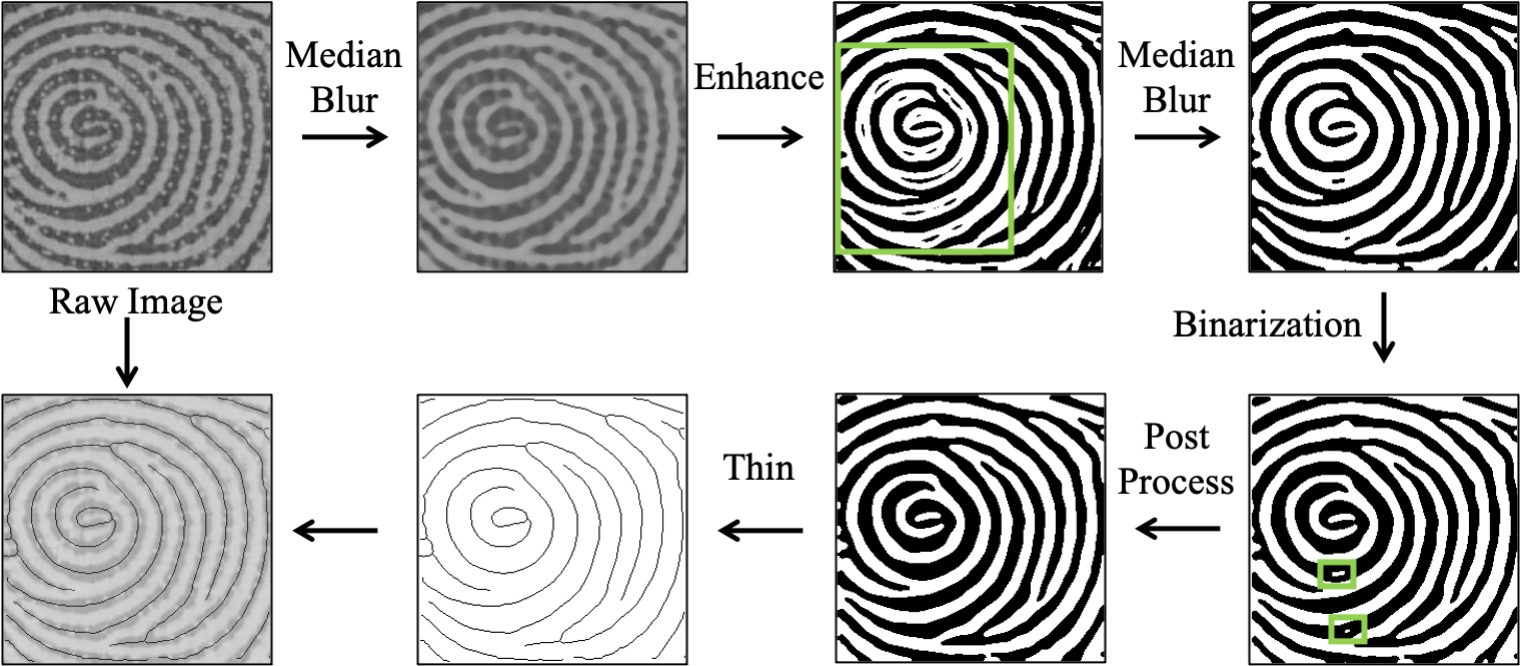}
  \caption{The binarization and thinning process of a fingerprint image.}
  \label{fig:binarization_thin}
  \Description{}
\end{figure}

\begin{figure}[t]
  \centering
  \includegraphics[width=0.7\linewidth]{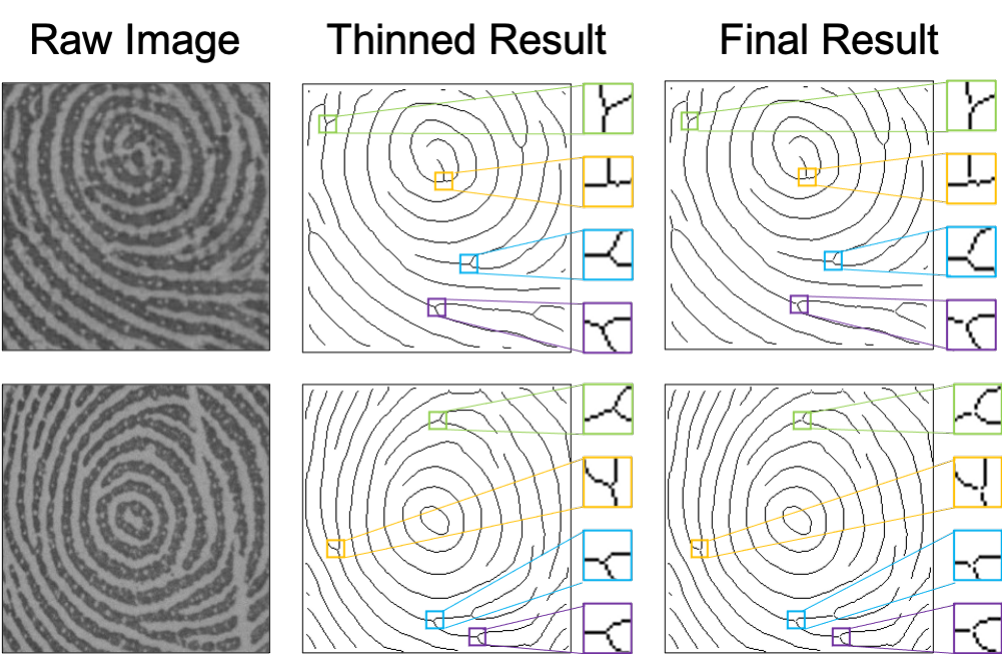}
  \caption{The results of removing Y-junctions of the ridges.}
  \label{fig:delete_y}
  \Description{}
\end{figure}

\begin{figure*}[t]
  \centering
  \includegraphics[width=0.9\linewidth]{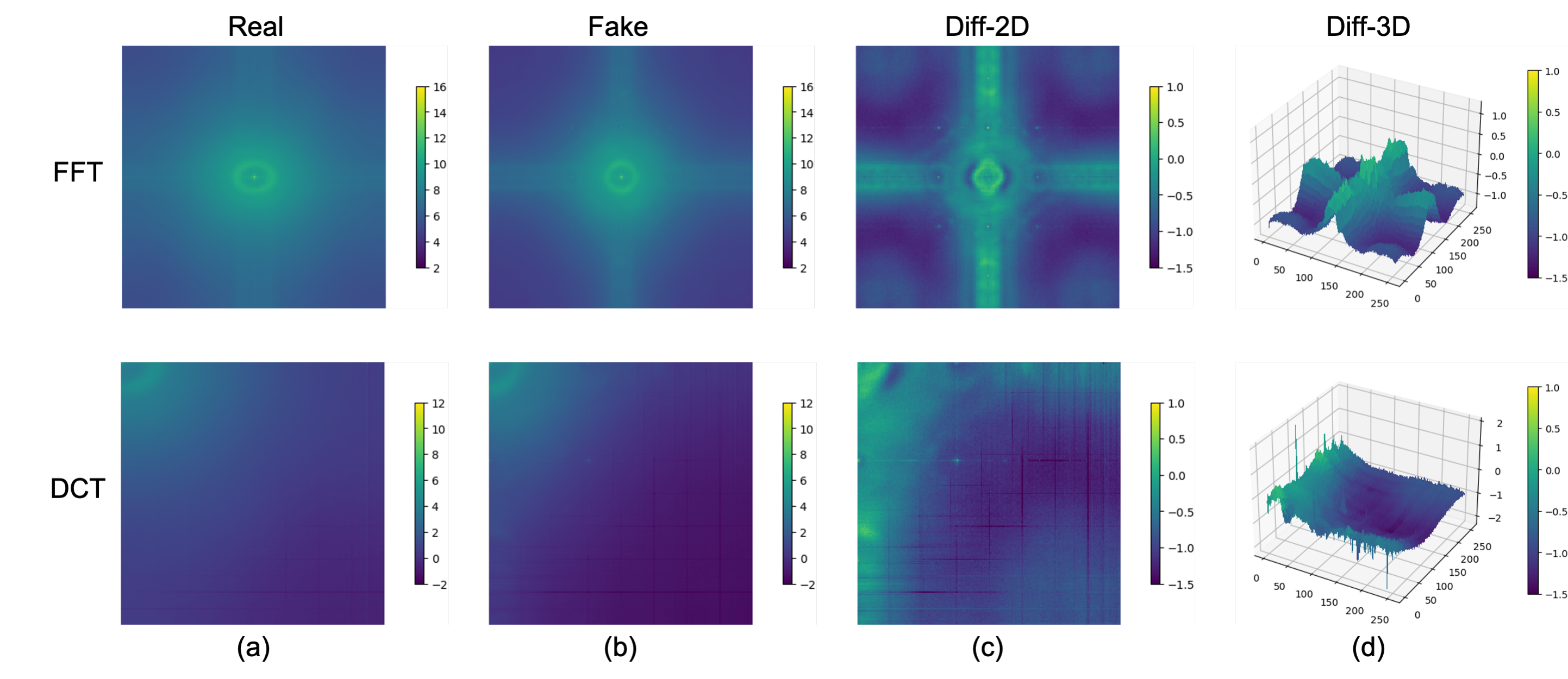}
  \caption{Frequency magnitude spectrums of the real and fake data, and the visualizations of their differences. The real and fake datasets are from the experimental data, which is described in Sec.~\ref{sec:experiment settings}.}
  \label{fig:freq_analysis}
  \Description{}
\end{figure*}

\subsection{Ridge Stream}
\label{sec:ridge stream}

Perspiration is a normal physiological phenomenon of live fingers. Sweat is generated from pores and can diffuse along the ridges, which can affect the grayscale values of the fingerprint images along ridges.
On the contrary, in fake fingerprint images, sweat pores are added to ridges according to a given distribution~\cite{wyzykowski2021level}. Besides, the grayscale variations along the ridges only rely on image translation methods (\textit{e.g.}, CycleGAN~\cite{zhu2017unpaired}) and is not processed separately, which cannot fully simulate realistic perspiration in GAN-generated fingerprint images.
Inspired by~\cite{derakhshani2003determination}, we exploit the statistical information along the ridges as the input of the ridge stream. 
Specifically, the ridge stream contains three steps:
1) image binarization and thinning, 2) feature extraction along the ridges, and 3) shallow neural network processing.

As shown in Fig.~\ref{fig:binarization_thin}, for the image binarization and thinning, the original fingerprint image is firstly filtered by a median filter to remove the pixels, which change significantly along the ridges (\textit{e.g.}, sweat pores).
This operation ensures the consistency of pixel values along the ridges and avoids ridge discontinuities in subsequent processing.
Then, we employ a Gabor filter, which utilizes the orientation and frequency information of the ridges, to further enhance the image.
Although we have employed a median filter to preprocess the original image, there still exists some errors in the enhanced results. 
Thus, the median filter is employed again to obtain a more accurate enhanced result.
Next, a threshold-based binarization method is utilized to binarize the enhanced image. 
Then, a post-processing algorithm, which fills the pores of the ridges, is performed, because these pores can cause wrong bifurcations in the thinning process.
Specifically, for each white pixel, its 24 neighboring pixels will be checked. If more than 15 neighboring pixels are black, the current pixel value is set to 0.
At last, the binarized ridges are thinned to single pixel width by using the image skeleton extraction algorithm~\cite{lee1994building}.

\begin{algorithm}[t]
  \DontPrintSemicolon
    
    \KwInput{$I$: A thinned fingerprint image with Y-junction removed ridges; $(p_x, p_y)$: Arbitrary point in the ridge;}
    \KwOutput{$curve$: Ordered pixel coordinates along the ridge where $(p_x, p_y)$ is located.}

    \SetKwFunction{FMain}{$findNextPoint$}
    \SetKwProg{Fn}{Function}{:}{}
    \Fn{\FMain{$I$, $p_x$, $p_y$, $curve$}}{
          $curve.append((p_x,p_y))$  \Comment{Add $(p_x,p_y)$ to curve; }\;
          $I[p_x][p_y] = 255$ \Comment{Remove $(p_x,p_y)$ in $I$; }\;
          $P = I[p_x-1:p_x+2][p_y-1:p_y+2]$\; \Comment{Get the local patch P of $(p_x,p_y)$;}\;
          $coord = argwhere(P == 0) - [1,1] + [p_x, p_y]$\; \Comment{Calculate the global coordinates $coord$ of black pixels within the local patch $P$;}\;
          \If{len($coord$) == 1}
          {
            $findNextPoint(I,coord[0][0], coord[0][1], curve)$\;
          }
          \ElseIf{len($coord$) == 2}
          {
            $curve_1, curve_2 = [], []$ \;
            $findNextPoint(I, coord[0][0], coord[0][1], curve_1)$\;
            $findNextPoint(I, coord[1][0], coord[1][1], curve_2)$\;
            $curve = curve_1.reverse + curve + curve_2$\;
          }
          \KwRet\ ;
    }

  \caption{Coordinate calculation of the ordered pixels}
  \label{algorithm:obtain_pixel}
  \end{algorithm}
  
    

  

Before extracting features along the ridges, the ridges with multiple bifurcations are required to be divided with a neighborhood-based method, to ensure each black pixel only appears on one ridge to avoid redundant calculations.
Specifically, for each black pixel, 8 neighboring pixels are checked. If there exists two or more potential directions within the local neighborhood, the current black pixel is set to white.
The Y-junction removing results are shown in Fig.~\ref{fig:delete_y}.
Then, we extract the individual ridges and obtain the pixel coordinates from the start point to the end point of each ridge with Algorithm~\ref{algorithm:obtain_pixel}.
Then, each ridge is further divided into segments, each of which contains 128 pixels, for convenience.
Next, as shown in Fig.~\ref{fig:pipeline}, we aggregate the pixel values along each ridge segment to form a 1D signal, which is represented as a curve. 
After a Gaussian smoothing operation, each 1D signal is processed by Fast Fourier Transform (FFT), and we select the average of these vectors $f_{raw}$ as the raw ridge feature, which can be formulated as
\begin{equation}
f_{raw}(\omega ) = \frac{\sum_{m=1}^{m=M}|\sum_{n=0}^{n=N-1} \mathsf{G_m} (n)e^{-j\frac{2\pi n\omega }{N} }| } {M},
\end{equation}
where $M$ represents the number of ridge segments, $N$ stands for the length of a ridge segment (\textit{e.g.}, 128), and $\mathsf{G_m}$ is the 1D signal of the m-th ridge.

At last, as shown in Fig.~\ref{fig:pipeline}, the raw features are further processed by a shallow neural network, which contains a batch normalization layer and a MLP, and can be defined as
\begin{equation}
  f_{ridge} = MLP(BatchNorm(f_{raw})),
\end{equation}
where $f_{ridge}$ is the final ridge feature obtained from the ridge stream.

\subsection{Generation Artifact Stream}
\label{sec:frequency stream}

Inspired by~\cite{frank2020leveraging}, which have revealed that there exists certain artifacts in the frequency spectrums of GAN-generated images, we firstly perform a frequency analysis for the real and GAN-generated fingerprint images. 
By comparing the discrepancy of the magnitude spectrums between the real and GAN-generated images with different frequency transforms (e.g. Fast Fourier Transform (FFT) and Discrete Cosine Transform (DCT)), we observe that the discrepancy between the FFT-based spectrums of the real and fake fingerprint images is relatively large.
As shown in Fig.~\ref{fig:freq_analysis}, we visualize the log-scale averaged magnitude spectrum after FFT and DCT over the real fingerprint dataset (Fig.~\ref{fig:freq_analysis} (a)) and fake fingerprint dataset (Fig.~\ref{fig:freq_analysis} (b)).
As can be observed, the FFT-based spectrum of the fake images, which are generated from the CycleGAN-based method~\cite{wyzykowski2021level}, possesses less energy at the high frequency locations, and the DCT-based spectrum of the fake images possesses severe grid-like patterns in the bottom-right corner.
To better explain these differences in the magnitude spectrum, we also plot the spectrum differences in 2D (Fig.~\ref{fig:freq_analysis} (c)) and 3D spaces (Fig.~\ref{fig:freq_analysis} (d)).
Apparently, the difference of the DCT spectrums is \textit{flatter} than that of the FFT spectrums, \textit{i.e.}, there exists larger differences and more severe artifacts in FFT-based spectrum.
Since the majority of the high frequency locations in the FFT-based spectrum of the fake images possess obviously less energy than the spectrum of real images, this observation is more stable than the grid-like artifacts in the DCT-based spectrums.
Therefore, based on the above analysis, the FFT magnitude spectrum is utilized as the input to a simple yet effective convolutional neural network to extract the generation artifact features.
As shown in Fig.~\ref{fig:pipeline}, we firstly transform the input images into the frequency domain via FFT, and compute its magnitude spectrum $f_{freq}$ as

\begin{equation}
  f_{freq}(x, y) = log[| \sum_{u=0}^{M-1}\sum_{v=0}^{N-1}I(u,v)e^{-j2\pi(\frac{ux}{M}+\frac{vy}{N}) } | + \epsilon],
  \label{equ:FFT2D}
\end{equation}
where $M$ and $N$ are the length and width of the image $I$, respectively. To avoid the value of 0, we set $\epsilon$ to $10^{-18}$.
Based on this 2D generation artifact feature $f_{freq}$, we design a simple yet effective convolutional neural network followed by a fully-connected network, to extract the generation artifact features, which can be described as:
\begin{equation}
  f_{artifact} = FC(Conv(f_{freq})),
\end{equation}
where $f_{artifact}$ is the final feature obtained from the generation artifact stream.

\subsection{Feature Fusion and Loss Function}
\label{sec:fusion_method}
As shown in Fig.~\ref{fig:pipeline}, the obtained features from the ridge stream ($f_{ridge}\in C$) and the generation artifact stream ($f_{artifact}\in C$) are fused in the fusion module.
Considering that the two streams are independent and their characteristics possess no similarity, we sum these two features directly, instead of utilizing the recently popular attention mechanism~\cite{luo2021generalizing}.
The fused features are fed into a MLP block to predict the authenticity of the input image, which can be described as
\begin{equation}
  Pred = MLP(f_{ridge} + f_{artifact}),
  \label{equ:feature_fusion}
\end{equation}
where $Pred$ is the output of the fusion module.

For the training of our proposed method, we adopt the cross-entropy loss as our loss function.

\section{Experiments}

\subsection{Experiment Settings}
\label{sec:experiment settings}

\subsubsection{\textbf{Datasets}}
To evaluate the effectiveness of our proposed method, we employ the Hong Kong Polytechnic University High Resolution Fingerprint database (PolyU-HRF)~\cite{zhao2009direct} and L3-SF Database~\cite{wyzykowski2021level} as the real and fake data, respectively.
The PolyU-HRF database consists of two sub-databases with different resolutions including 320*240 (DBI) and 640*480 (DBII).
To be consistent with the L3-SF database, we only utilize DBI as real data, including 1480 images captured from 148 fingers.
For the L3-SF database,~\cite{wyzykowski2021level} generates 5 sets of data via a CycleGAN-based method, each of which contains 1480 images generated from 148 masterprints.
To maintain the same diversity as PolyU-HRF database, the first set of L3-SF database is selected as the fake data.
In summary, the entire dataset employed in the experiments contains 2960 images, including 1480 real images from PolyU-HRF database and 1480 fake images from L3-SF database, and the resolution of these images is 320*240.
The entire dataset is divided into training set, validation set, and testing set. 
The training set contains 880 real images (from 88 fingers) and 880 fake images (from 88 masterprints).
For validation set and testing set, each of them contains 300 real images (from 30 fingers) and 300 fake images (from 30 masterprints).
For the convenience of using the anti-forensic method~\cite{dong2022think} to process the fake images, we center crop all the images to the resolution of 256*256.

\subsubsection{\textbf{Metrics}}
The traditional classification metrics, including accuracy and recall, are employed to evaluate the effectiveness and robustness in the experiments.
Besides, we compare the number of parameters possessed by different approaches to evaluate the complexity of these methods.

\begin{figure}[t]
  \centering
  \includegraphics[width=0.9\linewidth]{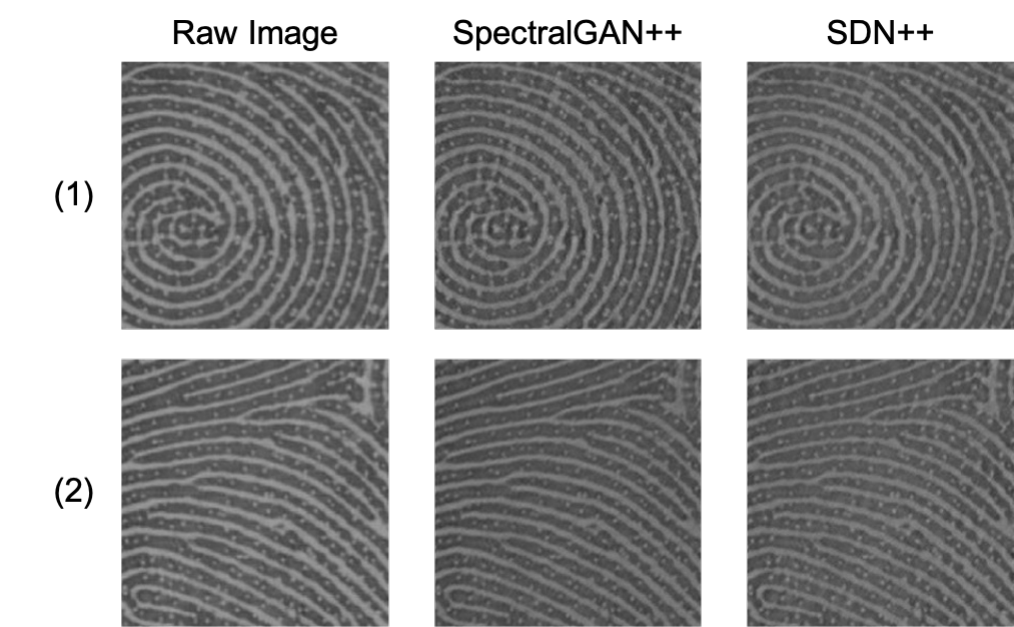}
  \caption{Two sets of fake fingerprint images before and after disturbed by~\cite{dong2022think}. The first column is the uncorrected fake images. The second and third columns are the disturbed images by employing SpectralGAN++ and SDN++, respectively.}
  \label{fig:attack_img}
  \Description{}
\end{figure}

\subsubsection{\textbf{Baseline methods}}
Since there is currently no deep forgery detection algorithm for fingerprint images, to the best of our knowledge, and the fake fingerprint images are generated via the CycleGAN-based method~\cite{wyzykowski2021level}, we select the state-of-the-art (SOTA) GAN-generated image detection methods (\textit{i.e.}, CNNDetection~\cite{wang2020cnn}, DCTAnalysis~\cite{frank2020leveraging}, LNP-based classifier~\cite{liu2022detecting}) as the baselines.
To ensure the fairness of the comparisons with these methods, we use the codes provided from the authors.

\subsubsection{\textbf{Anti-forensic Method}}
We also evaluate the robustness of the baseline and our methods against the anti-forensic method~\cite{dong2022think} in Sec.~\ref{sec:existing method} and Sec.~\ref{sec:ablation study}.
~\cite{dong2022think} firstly proposes SpectralGAN, which is developed based on CycleGAN~\cite{zhu2017unpaired} for unpaired spectrum-to-spectrum translation.
After processing the GAN-generated images with SpectralGAN, [11] further constructs a dictionary of power distribution for real data and performs a dictionary-based power distribution correction (PDC) to mitigate the spectral artifacts.
Besides,~\cite{dong2022think} presents another method, which introduces spectrum difference normalization (SDN) by utilizing the averaged spectrum differences between the fake and real images to correct the spectrums. 
After magnitude spectrum correction, it again performs a dictionary-based power distribution correction (PDC) to mitigate the spectral artifacts.
Note that the combination of SpectralGAN and PDC is referred to as SpectralGAN++, and the combination of SDN and PDC is referred to as SDN++, in this paper.

\begin{figure}[t]
  \centering
  \includegraphics[width=1.0\linewidth]{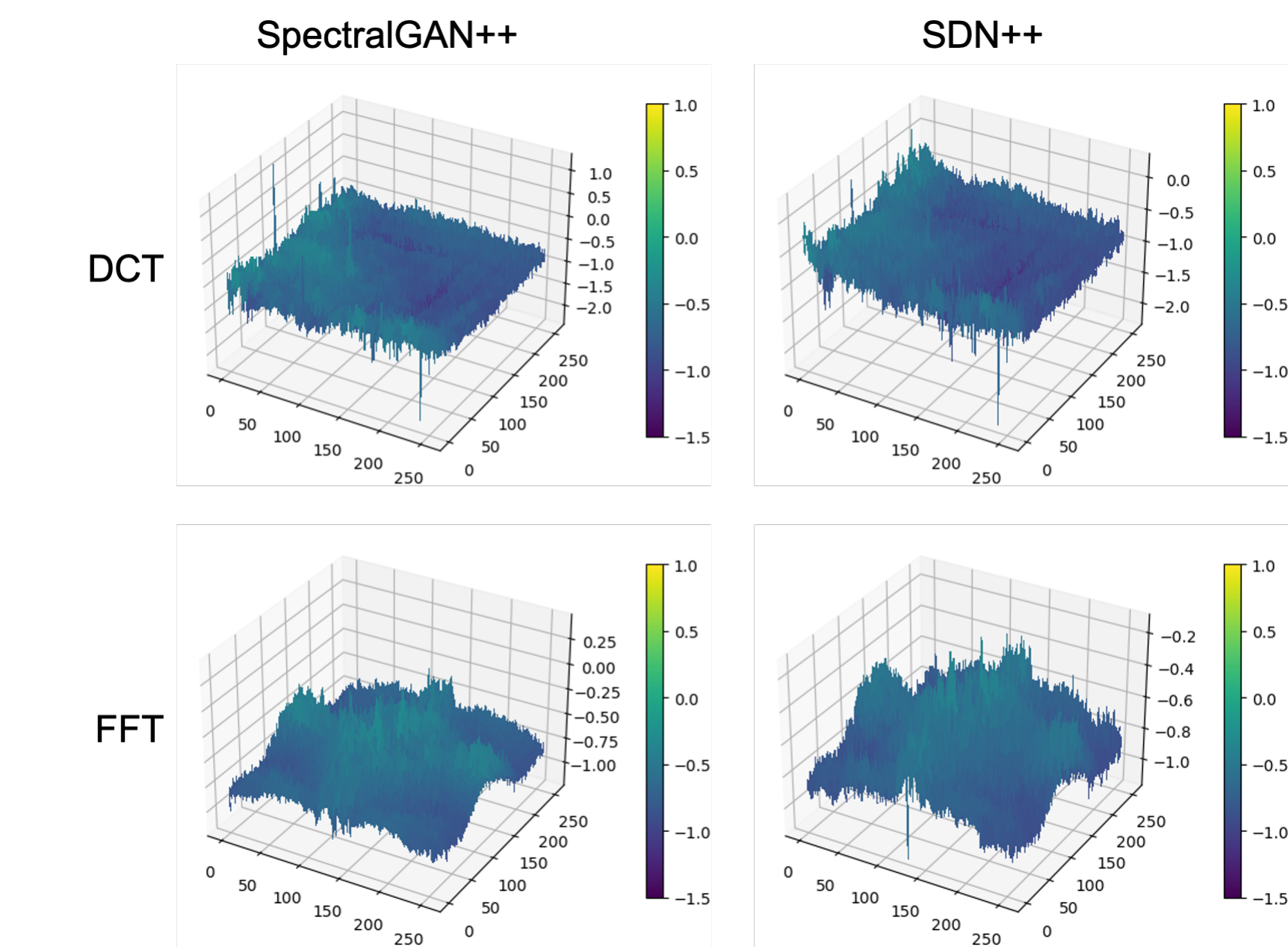}
  \caption{Visualizations of the spectral differences between the real and corrected fake data.}
  \label{fig:attack_freq}
  \Description{}
\end{figure}

\subsection{Implementation Details}
\label{sec:implementation}
To avoid the inconsistencies induced by different image formats, we firstly unify the formats of the fake data into the JPEG format, which is consistent with the real data.
For the ridge stream, we use the Gabor filter in fingerprint-enhancer~\cite{fingerprintenhancer}, which is a python code library, to enhance the fingerprint images.
The threshold of the binarization step is set to 100. The MLP block in the ridge stream consists of two fully-connected layers with batch normalization, relu, and dropout layers.
The output dimension of this stream (\textit{i.e.}, $C$) is set to 128.
For the generation artifact stream, the constructed shallow neural network consists of two convolutional layers with batch normalization, relu, and maxpooling layers, followed by an adaptive maxpooling layer and a fully-connected layer.
Xavier~\cite{glorot2010understanding} initialization method is employed to initialize the parameters of the entire method.
Adam~\cite{kingma2014adam} optimizer is utilized for optimization, and the initial learning rate is set to 0.001 with a CosineAnnealingLR scheduler~\cite{loshchilov2016sgdr} being exploited.
In the training process, the weight decay is set to 0.0001, and the batch size is set to 32. 
Since the two streams are independent, we randomly flip the input in each stream during the training process to further improve the robustness. 
The early stopping strategy is utilized to stop the training process, when the accuracy of the validation set does not increase for 5 consecutive epochs.

To better reproduce the results of the anti-forensic method~\cite{dong2022think}, \textit{i.e.}, SpectralGAN++ and SDN++, we employ the official code of SpectralGAN provided by the authors.
Since the source codes of SDN and PDC are not provided, we implement them by following the detailed instructions in the original paper.
Our reproduction results are shown in Fig.~\ref{fig:attack_img}. 
To demonstrate that the reproduced method can disturb the magnitude spectrums of the fake images, we visualize the spectral differences between the disturbed and real data in Fig.~\ref{fig:attack_freq}.
Compared to Fig.~\ref{fig:freq_analysis} (d), we can observe that the energy drop in the high frequency locations has been alleviated in the FFT spectrums and the differences of the DCT spectrums become minor, which proves that the generation artifacts are indeed mitigated after the correction.

\begin{table*}[t]
  \caption{Comparisons with the SOTA methods on original testing set and disturbed testing set. (For each column, the highest value is highlighted in bold font, and the symbol $^*$ denotes the second best value.)}
  \label{tab:existing_method_total}
  \renewcommand\arraystretch{1.1}
  \tabcolsep=8pt
  \begin{tabular}{r|c|c|c|c|c|c|c}
    \hline \hline 
    \multirow{3}{*}{Method} & \multirow{3}{*}{Params} & \multicolumn{2}{c|}{\multirow{2}{*}{Original Data}} &  \multicolumn{4}{c}{Corrected Data} \\
    \cline{5-8}
    & & \multicolumn{2}{c|}{}  & \multicolumn{2}{c|}{SpectralGAN++} & \multicolumn{2}{c}{SDN++} \\
    \cline{3-8}
    & & Recall & Acc & Recall & Acc & Recall & Acc \\
    \hline
    CNNDet (pretrained)~\cite{wang2020cnn} & 23.510M & 24.00 & 58.17 & 10.07 & 51.34 & 2.33 & 47.33 \\
    CNNDet (B+J 0.1)~\cite{wang2020cnn} & 23.510M & 100.00 & 100.00 & 53.69 & 76.92 & 74.00 & 87.00 \\
    CNNDet (B+J 0.5)~\cite{wang2020cnn} & 23.510M & 100.00 & 100.00 & $76.17^*$ & $88.13^*$ & 75.00 & 87.50  \\
    LNP-based Classifier~\cite{liu2022detecting} & 23.516M & 100.00 & 100.00 & 64.09 & 82.10 & $92.67^*$ & $96.33^*$ \\
    DCTAnalysis~\cite{frank2020leveraging} & \textbf{38.831K} & 98.28 & 99.13 & 60.14 & 80.90 & 80.07 & 90.45 \\
    Ours & $94.786\mathrm{K}^*$ & \textbf{100.00} & \textbf{100.00} & \textbf{92.28} & \textbf{96.15} & \textbf{99.00} & \textbf{99.50} \\
    \hline \hline 
\end{tabular}
\end{table*}

\begin{figure}[t]
  \centering
  \includegraphics[width=0.8\linewidth]{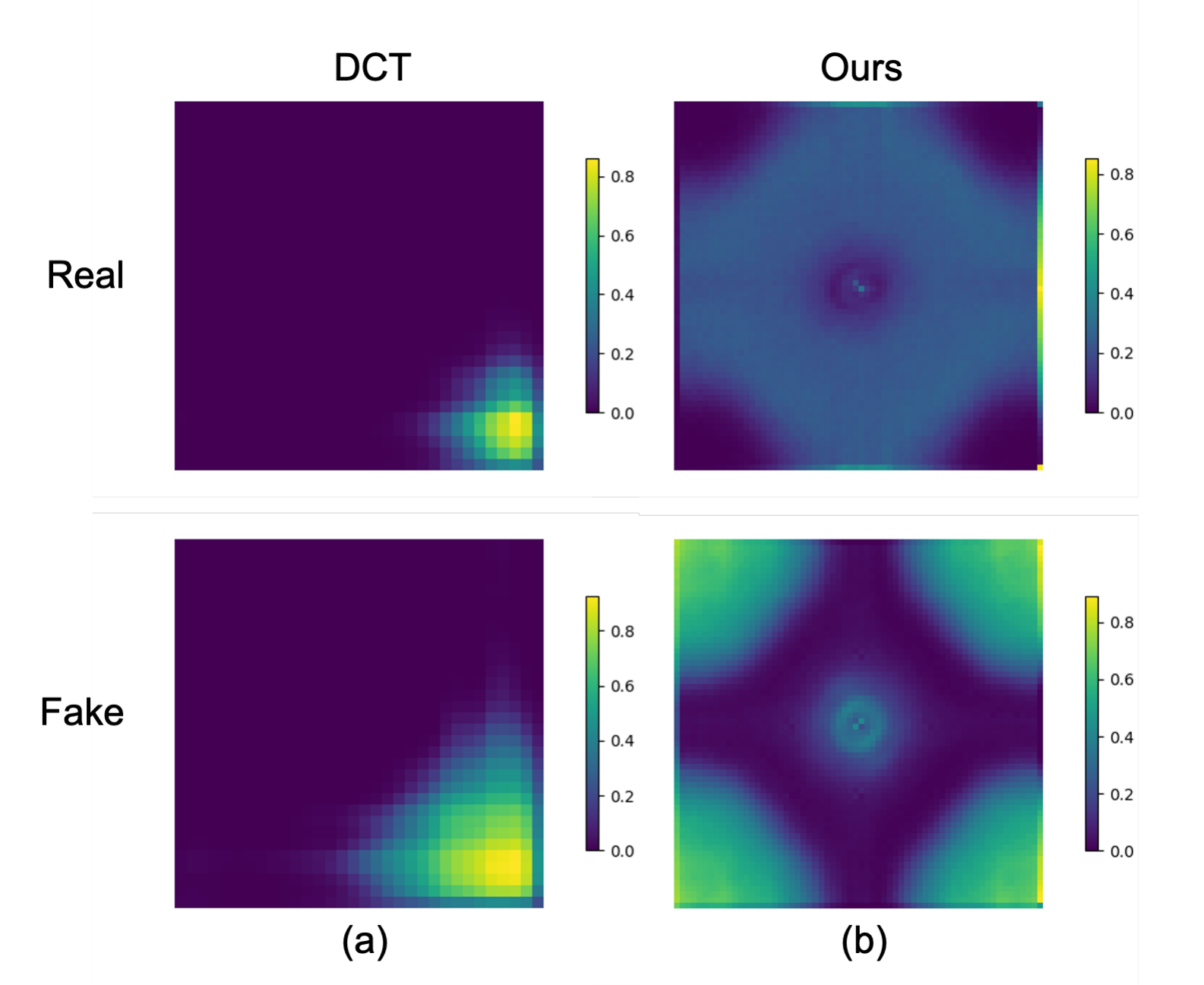}
  \caption{Visualizations of the activation maps of the top convolutional layer in DCTAnalysis and our generation artifact stream for real and fake images.}
  \label{fig:cam}
  \Description{}
\end{figure}

\subsection{Performance Comparisons}
\label{sec:existing method}

\subsubsection{\textbf{Effectiveness and complexity}}
In this experiment, we compare our method with the existing methods on the original undisturbed data.
As shown in Tab.~\ref{tab:existing_method_total} (the Original Data column), our approach can correctly distinguish generated data from real data (\textit{i.e.}, recall = 100.00\%, acc = 100.00\%).
For CNNDetection~\cite{wang2020cnn}, when we directly employ the provided pre-trained model in the experiment, the performance is extremely poor (\textit{i.e.}, recall = 24.00\%, acc = 58.17\%).
Thus, we retrain this network by using the training set of our method and the data augmentation strategy\&settings in~\cite{wang2020cnn}.
It is obvious that the performance of CNNDetection~\cite{wang2020cnn} is significantly improved after retraining, \textit{i.e.}, both the accuracy and recall reach 100.00\%.
This phenomenon also indicates that the existing universal deep forgery detectors cannot be directly generalized to fingerprint images, i.e., they must be retrained to achieve good performance.
For the LNP-based classifier~\cite{liu2022detecting}, since the authors have not provided the pretrained model, we directly retrain this network with the authors' code. 
Its results reveal that the fake data can also be completely distinguished from the real data (\textit{i.e.}, acc = 100.00\%).
Although CNNDetection~\cite{wang2020cnn} and the LNP-based classifier~\cite{liu2022detecting} can achieve good performance, the backbone networks of these two methods are both ResNet-50, which possesses a large number of parameters ($\thickapprox$23.5M) compared to our method ($\thickapprox$94.8K).
Apparently, our method can achieve the SOTA performance with a much lower complexity.

For DCTAnalysis~\cite{frank2020leveraging}, the results in Tab.~\ref{tab:existing_method_total} (the Original Data column) show that our method outperforms DCTAnalysis on recall by 1.72\%. 
Meanwhile, DCTAnalysis constructs a simple convolutional neural network with very few parameters.
However, since our method extracts both the ridge and generation artifact information, the number of parameters of our method is slightly larger than DCTAnalysis.

To better illustrate the superiority of our FFT-based features, Grad-CAM~\cite{selvaraju2017grad} is utilized to visualize the activation maps of the top convolutional layer of DCTAnalysis and our generation artifact stream for real and fake images in Fig.~\ref{fig:cam}.
As can be observed, DCTAnalysis actually focuses on the existence of the grid-like artifacts, which are usually located in the bottom-right corner region.
On the contrary, our FFT-based method focuses on the entire energy concertrations, which will be disturbed less as proved in Fig.~\ref{fig:attack_freq}. Specifically, for the real data, energy is mainly concentrated in the central regions, while a portion of the energy still exists in the high frequency locations, which matches the corresponding activation map. 
On the contrary, for the fake data, the energy is much less in the high frequency regions. Therefore, our FFT-based method is more robust than the traditional DCT-based method.

\begin{table*}[t]
  \caption{Performances of each stream and different fusion methods on original testing set and disturbed testing set. (For each column, the highest value is highlighted in bold font, and the symbol $^*$ denotes the second best value.)}
  \label{tab:ablation_study}
  \renewcommand\arraystretch{1.1}
  \tabcolsep=8pt
  \begin{tabular}{cc|cc|c|c|c|c|c|c}
    \hline \hline 
    \multirow{3}{*}{\makecell{Ridge \\ Stream }} & \multirow{3}{*}{\makecell{Generation\\ Artifact }} & \multicolumn{2}{c|}{\multirow{2}{*}{Fusion Method}}  & \multicolumn{2}{c|}{\multirow{2}{*}{Original Data}} &  \multicolumn{4}{c}{Corrected Data} \\
    \cline{7-10}
    &  & \multicolumn{2}{c|}{} &  \multicolumn{2}{c|}{}  & \multicolumn{2}{c|}{SpectralGAN++} & \multicolumn{2}{c}{SDN++} \\
    \cline{3-10}
    &  & concat & sum & Recall & Acc & Recall & Acc & Recall & Acc \\
    \hline
    \usym{1F5F8} &  & - & - & 93.33 & 91.67 & 79.87 & 84.95 & 80.33 & 85.17 \\
     & \usym{1F5F8} & - & - & 100.00 & 100.00 & 86.91 & 93.48 & 99.00 & 99.50  \\
    \hline
    \usym{1F5F8} & \usym{1F5F8} & \usym{1F5F8} &  & 100.00 & 100.00 & $89.26^*$ & $94.65^*$ & 99.00 & 99.50 \\
    \usym{1F5F8} & \usym{1F5F8} &  & \usym{1F5F8} & \textbf{100.00} & \textbf{100.00} & \textbf{92.28} & \textbf{96.15} & \textbf{99.00} & \textbf{99.50} \\
    \hline \hline 
\end{tabular}
\end{table*}

\subsubsection{\textbf{Robustness}}
We regard SpectralGAN++ and SDN++ as the anti-forensic methods to illustrate the robustness of each forgery detection method against these attacks.
As shown in Tab.~\ref{tab:existing_method_total}, the performances of all the baseline methods are significantly decreased, when the testing set is disturbed by the anti-forensic methods.
It is interesting that the performance of CNNDetection, which only extracts features in spatial domain, is also decreased. If a larger proportion of data augmentations is employed, CNNDetection becomes more robust (\textit{i.e.}, acc=88.13\% \textit{vs.} acc=76.92\%, acc=87.50\% \textit{vs.} acc=87.00\%).
Therefore, we conclude that CNNDetection actually implicitly learns the invisible frequency-domain artifacts, which can be easily modified by the anti-forensic methods~\cite{dong2022think}.
Since the LNP-based classifier utilizes the high-frequency artifacts and DCTAnalysis focuses on the DCT magnitude spectrum, both of their utilized features are easily suppressed by~\cite{dong2022think}.
Therefore, the performances of these methods are significantly degraded.
Although the magnitude spectrum of FFT is also mitigated, since our method not only focuses on a more robust frequency domain feature, \textit{i.e.}, the trend of energy concentrations, but also exploits the fingerprint-specific ridge feature, our method possess better robustness against these anti-forensic attacks.

\subsection{Ablation Study}
\label{sec:ablation study}

For fair comparisons, we directly add a fully-connected layer after the top layer of each stream to output the prediction results of each single stream.
We firstly demonstrate the effectiveness of each stream in Tab.~\ref{tab:ablation_study} (the Original Data column). 
As can be observed, the generation artifact stream can correctly distinguish the fake data from the real data (\textit{i.e.}, acc=100.00\%).
Since our ridge stream only extracts the features from the ridges, and this stream has not specifically learned the 2D frequency information, its performance is slightly poor compared to the generation artifact stream.
Although this stream cannot achieve perfect performance, the introduction of this stream can effectively avoid the overfitting to certain frequency information.

In Tab.~\ref{tab:ablation_study} (the Corrected Data column), we also quantitatively analyze the robustness of each stream.
As can be observed, the performance of each stream is obviously decreased against the attack of SpectralGAN++. However, the robustness is improved after fusing the two streams.
Since the artifact mitigation performance of SDN++ are not as good as SpectralGAN++, as shown in Fig. 7, all the methods against SDN++ can achieve better results compared to these against SpectralGAN++. Besides, since our FFT-based feature focuses on the entire energy concertrations rather than certain specific artifacts, our generation artifact stream can already achieve an excellent result against SDN++. Thus, our method does not give further performance gain against SDN++ when the ridge stream is combined with the generation artifact stream.

In Tab.~\ref{tab:ablation_study}, we also evaluate different feature fusion methods of the two streams. 
As can be observed, the robustness of \textit{sum} is better than \textit{concat}.
In general, the above results have demonstrated that the combination of the ridge stream and generation artifact stream can improve the overall effectiveness and robustness against the anti-forensic methods~\cite{dong2022think}.

\section{Conclusion}
In this paper, we propose the first deep forgery detection approach for fingerprint images, to the best of our knowledge.
Instead of directly utilizing the original images or their frequency spectrums for feature extraction, we introduce a unique ridge stream, which is an unique characteristic of the fingerprint images, extract features from the grayscale variations along the ridges.
Then, we construct a generation artifact stream, which exploits the generation artifacts of the GAN-generated images, by extracting the generation artifact features from the FFT frequency spectrum.
At last, we sum and process the features of these two streams for the final prediction. Note that the complexity of each stream in our method are small due to our lightweight designs.
Comprehensive experiments have demonstrated that our proposed approach is effective, robust to the anti-forensic method~\cite{dong2022think}, with low complexity, compared to the existing methods.

\bibliographystyle{ACM-Reference-Format}
\bibliography{RFDforFin}

\end{sloppypar}
\end{document}